\pdfoutput=1

\documentclass[11pt]{article}

\usepackage{acl}

\usepackage{times}
\usepackage{latexsym}

\usepackage[T1]{fontenc}

\usepackage[utf8]{inputenc}

\usepackage{microtype}

\usepackage{inconsolata}

\usepackage{graphicx}
\usepackage{microtype}
\usepackage{bbold}
\usepackage{multirow}
\usepackage{booktabs}
\usepackage{xcolor}
\usepackage{colortbl}
\usepackage{hhline}
\usepackage{pifont}
\usepackage{amsmath}
\definecolor{darkgreen}{HTML}{228B22}
\definecolor{nblue}{HTML}{377eb8}

\definecolor{ggreen}{HTML}{1b9e77}
\definecolor{oorange}{HTML}{d95f02}
\definecolor{bblue}{HTML}{7570b3}
\definecolor{ppurple}{HTML}{e37fbb}

\definecolor{lgreen}{HTML}{9CD24A}
\definecolor{yyellow}{HTML}{FFD52D}
\definecolor{ggold}{HTML}{E1BC89}
\definecolor{ggray}{HTML}{AAAAAA}

\newcommand{\xmark}{\textcolor{red}{\ding{55}}}
\newcommand{\cmark}{\textcolor{darkgreen}{\ding{51}}}

\newcommand{\colorhl}[2]{\colorbox{#1!20}{{#2}}}

\title{AutoTemplate: A Simple Recipe for Lexically Constrained Text Generation}

\author{Hayate Iso \\
  Megagon Labs \\
  \texttt{hayate@megagon.ai}}

\begin{document}
\maketitle
\begin{abstract}
Lexically constrained text generation is one of the constrained text generation tasks, which aims to generate text that covers all the given constraint lexicons. While the existing approaches tackle this problem using a lexically constrained beam search algorithm or dedicated model using non-autoregressive decoding, there is a trade-off between the generated text quality and the hard constraint satisfaction.
We introduce AutoTemplate, a simple yet effective lexically constrained text generation framework divided into template generation and lexicalization tasks.
The template generation is to generate the text with the placeholders, and lexicalization replaces them into the constraint lexicons to perform lexically constrained text generation.
We conducted the experiments on two tasks: keywords-to-sentence generations and entity-guided summarization.
Experimental results show that the AutoTemplate outperforms the competitive baselines on both tasks while satisfying the hard lexical constraints.\footnote{The code is available at \url{https://github.com/megagonlabs/autotemplate}}
\end{abstract}

\section{Introduction}

Text generation often requires lexical constraints, i.e., generating a text containing pre-specified lexicons.
For example, the summarization task may require the generation of summaries that include specific people and places~\cite{fan-etal-2018-controllable,he-etal-2022-ctrlsum}, and advertising text requires the inclusion of pre-specified keywords~\cite{miao2019cgmh,zhang-etal-2020-pointer}.

\begin{figure}[t]
    \centering
    \includegraphics[width=\linewidth]{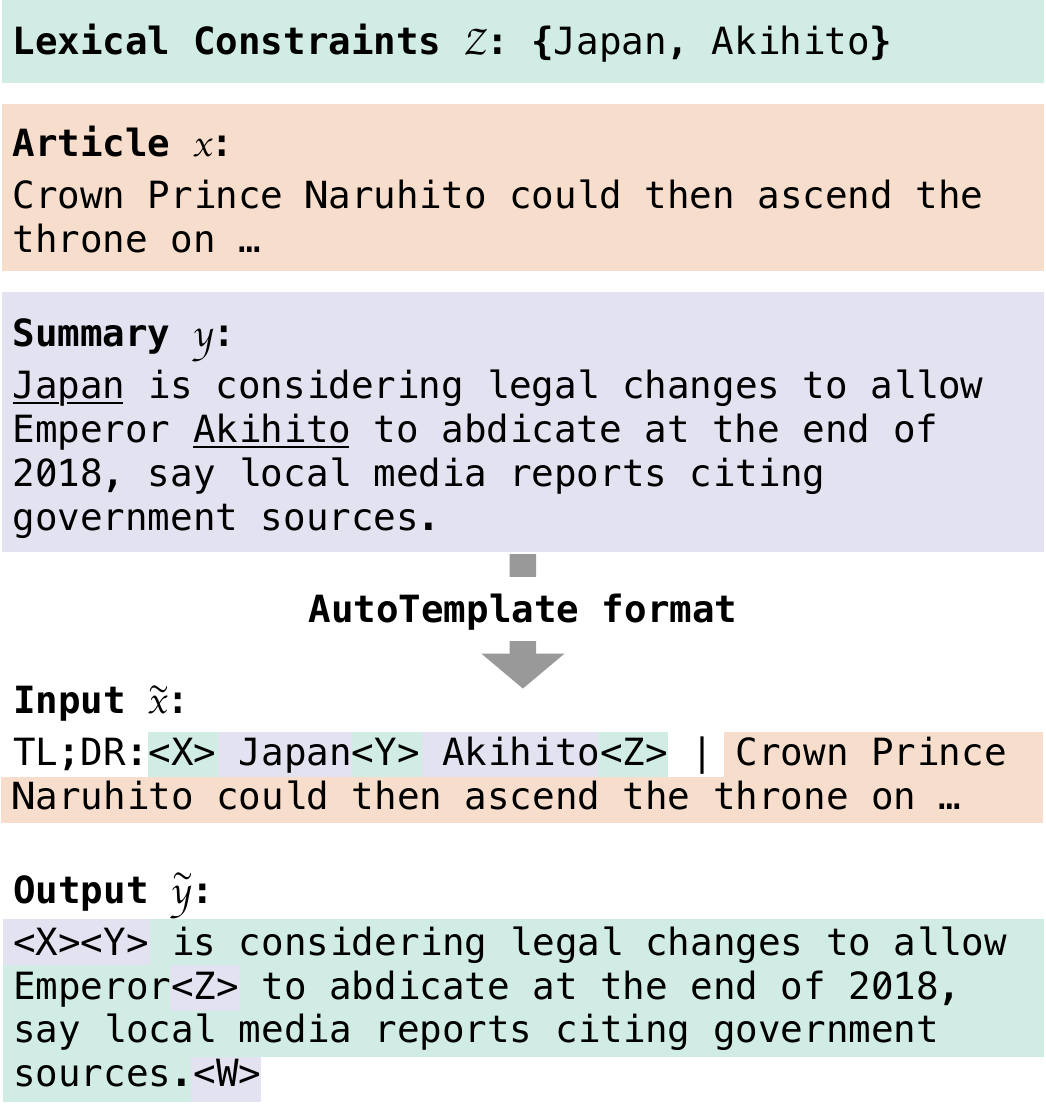}
    \caption{Illustration of AutoTemplate. We build the model input $\tilde{x}$ by concatenating the constraint lexicons $\mathcal{Z}$ with mask tokens. For the conditional text generation task, we further concatenate input document $x$. We also build the model output $\tilde{y}$ by masking the constraint lexicons in summary $y$. Then, we can train a standard sequence-to-sequence model, $p(\tilde{y} \mid \tilde{x})$, generate masked template $\tilde{y}$ given input $\tilde{x}$, and post-process to achieve lexically constrained text generation.}
    \label{fig:overview}
\end{figure}

However, the black-box nature of recent text generation models with pre-trained language models~\cite{devlin-etal-2019-bert,brown2020gpt3} makes it challenging to impose such constraints to manipulate the output text explicitly.
\citet{hokamp-liu-2017-lexically} and others tweaked the beam search algorithm to meet lexical constraints by increasing the weights for the constraint lexicons, but it often misses to include all the constrained lexicons.
\citet{miao2019cgmh} and others introduced specialized non-autoregressive models~\cite{gu2018nonautoregressive} that insert words between the constraint lexicons, but the generated texts tend to be lower-quality than standard autoregressive models.

On the other hand, classical template-based methods~\cite{kukich-1983-design} can easily produce text that satisfies the lexical constraints as long as we can provide appropriate templates.
Nevertheless, it is impractical to prepare such templates for every combination of constraint lexicons unless for specific text generation tasks where the output text patterns are limited, such as data-to-text generation tasks~\cite{angeli-etal-2010-simple}. Still, if such a template could be \textit{generated automatically}, it would be easier to perform lexically constrained text generation.

We propose AutoTemplate, a simple framework for lexically constrained text generations by automatically generating templates given constrained lexicons and replacing placeholders in the templates with constrained lexicons.
The AutoTemplate, for example, can be used for summarization tasks, as illustrated in Figure~\ref{fig:overview}, by replacing the constraint lexicons (i.e., $\{\text{Japan, Akihito}\}$) in the output text with placeholder tokens during training and using these constraints as a prefix of the input, creating input-output pairs, and then using a standard auto-regressive encoder-decoder model~\cite{ilya2014seq} to train the AutoTemplate model.
During the inference, the constraint lexicons are prefixed in the same way, the model generates the template for the constraints, and the placeholder tokens are replaced with the constraint lexicons to perform lexically constrained text generation.

We evaluate AutoTemplate across two tasks: keywords-to-sentence generation on One-Billion-Words and Yelp datasets (\S\ref{sub:k2sg}), and entity-guided summarization on CNNDM~\cite{hermann2015teaching} and XSum datasets~\cite{narayan-etal-2018-dont}
(\S\ref{sub:entity_guilded}).
The AutoTemplate shows better keywords-to-sentence generation and entity-guided summarization performance than competitive baselines, including autoregressive and non-autoregressive models, while satisfying hard lexical constraints.
We will release our implementation of AutoTemplate under a BSD license upon acceptance.

\begin{table*}[t]
    \centering
    \scriptsize
    \resizebox{\textwidth}{!}{
    \begin{tabular}{l|cccccc}
        \toprule
        & multiple keywords & autoregressive decoding& keyword conditioning & constraint satisfaction \\\midrule
        SeqBF~\scriptsize{\cite{mou-etal-2016-sequence}} & \xmark & \xmark & \cmark & \cmark\\
        CGMH~\scriptsize{\cite{miao2019cgmh}} & \cmark & \xmark & \cmark & \cmark\\
        GBS~\scriptsize{\cite{hokamp-liu-2017-lexically}} & \cmark & \cmark & \xmark & \xmark \\
        CTRLsum~\scriptsize{\cite{he-etal-2022-ctrlsum}} & \cmark & \cmark & \cmark & \xmark\\
        InstructGPT~\scriptsize{\cite{ouyang2022training}} & \cmark & \cmark & \cmark & \xmark\\
        \midrule
        AutoTemplate (ours) & \cmark & \cmark & \cmark & \cmark\\
        \bottomrule
    \end{tabular}
    }
    \caption{Summary of existing work for lexically constrained text generation. SeqBF~\cite{mou-etal-2016-sequence} and CGMH~\cite{miao2019cgmh} use non-autoregressive decoding methods to insert words between given keywords. While these methods easily satisfy the lexical constraints, in general, non-autoregressive methods tend to produce lower-quality text generation than autoregressive methods. GBS~\cite{hokamp-liu-2017-lexically}, CTRLSum~\cite{he-etal-2022-ctrlsum}, and InstructGPT~\cite{ouyang2022training} use autoregressive methods to perform text generation, but there is no guarantee to satisfy all lexical constraints. AutoTemplate empirically demonstrates the capability to generate text that satisfies the constraints.}
    \label{tab:related}
\end{table*}

\section{AutoTemplate}
\label{sec:method}
AutoTemplate is a simple framework for lexically constrained text generation (\S\ref{sub:def}), divided into two steps: template generation (\S\ref{sub:template}) and lexicalization (\S\ref{sub:inference}).
The template generation task aims to generate the text with placeholders $\tilde{y}$, which we defined as a template, given constraint lexicons $\mathcal{Z}$, and the lexicalization is to replace these placeholders with the constraints to perform lexically constrained text generation.

\subsection{Problem Definition}
\label{sub:def}
Let $x$ be a raw input text, and $\mathcal{Z}$ be a set of constraint lexicons; the goal of the lexically constrained text generation is to generate a text $y$ that includes all the constraint lexicons $\mathcal{Z}$ based on the input text $x$.
For example, given a news article $x$ and some entities of interest $\mathcal{Z}$, the task is to generate a summary $y$ that includes all entities.
Note that unconditional text generation tasks, such as keywords-to-sentence generation (\S\ref{sub:k2sg}), are only conditioned by a set of lexicons $\mathcal{Z}$, and in this case, we treat the input data $x$ as empty to provide a unified description without loss of generality.

\subsection{Template Generation}
\label{sub:template}
Given training input-output pairs $(x, y)$ and constraint lexicons $\mathcal{Z}$, we aim to build a model that generates a template $\tilde{y}$, which has the same number of placeholder tokens as the constraint lexicons $\mathcal{Z}$.
We assume that the output text $y$ in the training set includes all the constraint lexicons $\mathcal{Z}$.

The template $\tilde{y}$ is created by replacing the constraint lexicon $\mathcal{Z}$ in the output text $y$ with unique placeholder tokens according to the order of appearances (i.e., \texttt{<X>}, \texttt{<Y>}, and \texttt{<Z>} in Figure~\ref{fig:overview}),\footnote{We also prefix and postfix the placeholder tokens to use them as BOS and EOS tokens.} and then the model input $\tilde{x}$ is created by prefixing the constraint lexicons $\mathcal{Z}$ with the raw input text $x$.\footnote{We use $|$ as separator token for constraints $\mathcal{Z}$ and input text $x$ and also prefixed \texttt{TL;DR:}.} These lexicons $\mathcal{Z}$ are concatenated with the unique placeholder tokens to let the model know the alignment between input and output. We discuss this design choice in \S\ref{sec:analysis}.

Using the AutoTemplate input-output pairs $(\tilde{x}, \tilde{y})$, we can build an automatic template generation model $p(\tilde{y} | \tilde{x})$ using any sequence-to-sequence models.
This study builds the template generation model $p$ using an autoregressive Transformer model with a regular beam search~\cite{vaswani2017transformer}.

\subsection{Lexicalization}
\label{sub:inference}
After generating the template $\tilde{y}$, we replace the placeholder tokens with constraint lexicons $\mathcal{Z}$ as post-processing to achieve lexically constrained text generation.
Specifically, during inference, constraint lexicons are prefixed to the input text $x$ in the same way to build the model input $\tilde{x}$.
Then, we can obtain the template $\tilde{y}$ from the model $p$ and replace the placeholder tokens with the constraint lexicons $\mathcal{Z}$.

\subsection{Comparison with existing approaches}
An important contribution of this study is to show that lexically-constrained generation can be performed in a simple way with AutoTemplate, whereas it was previously done with only complicated methods.
As summarized in Table~\ref{tab:related}, SeqBF~\cite{mou-etal-2016-sequence} is the first neural text generation model for lexically constrained text generation based on non-autoregressive decoding. The SeqBF performs lexically constrained text generation by generating forward and backward text for a given constraint lexicon.
The most significant limitation is that only a single keyword can be used for the constraint.

CGMH~\cite{miao2019cgmh} and similar models~\cite{zhang-etal-2020-pointer,he-2021-parallel} are yet another non-autoregressive models that achieve lexicon-constrained generation by inserting words between given constraint vocabularies, thus easily incorporating multiple constraints into the output text.
Nevertheless, non-autoregressive models require complicated modeling and training to generate text as good as that of autoregressive models.
We confirmed that the AutoTemplate produces consistently higher quality text than non-autoregressive methods, with or without leveraging pre-training (\S\ref{sub:k2sg}).

Another direction is to incorporate \textit{soft} constraints into the autoregressive models such as constrained beam search~\cite{hokamp-liu-2017-lexically,post-vilar-2018-fast} and keywords conditioning~\cite{he-etal-2022-ctrlsum}.
GBS~\cite{hokamp-liu-2017-lexically} is a constrained bean search technique that incorporates multiple keywords as constraints and promotes the inclusion of those keywords in the output during beam search. However, GBS often misses keywords in the output text.

CTRLSum~\cite{he-etal-2022-ctrlsum} imposes keyword conditioning into encoder-decoder models by prefixing the keywords with the input.
This method can be easily conditioned with multiple keywords as a prefix and can be implemented on an autoregressive model, resulting in high-quality text generation.
However, the CTRLSum model cannot guarantee to satisfy lexical constraints.
Our experiments show that as the number of constraints increases, it is more likely to miss constraint lexicons in the output text (\S\ref{sub:entity_guilded}).

InstructGPT~\cite{ouyang2022training} has shown remarkable zero-shot ability in many NLP tasks, and lexically constrained text generation is no exception.
Our experiments confirmed that the model can generate a very fluent sentence, but as with CTRLSum, we observed a significant drop in the success rate with each increase in the number of keywords.\footnote{Recent studies have pointed out that ambiguity in instructions influences output quality, but this issue remains to be addressed in future work~\cite{zhang-etal-2024-xatu,niwa2024ambignlgaddressingtaskambiguity}.}
\begin{table*}[t]
    \centering
    \scriptsize
    \resizebox{2\columnwidth}{!}{
    \begin{tabular}{l|cccccc|cccccc}
        \toprule
        \multirow{2}{*}{\textbf{Model}} & \multicolumn{6}{c}{\textbf{One-Billion-Word}} & \multicolumn{6}{c}{\textbf{Yelp}}\\
        & \textbf{B2} & \textbf{B4} & \textbf{N2} & \textbf{N4} & \textbf{M} & \textbf{SR} & \textbf{B2} & \textbf{B4} & \textbf{N2} & \textbf{N4} &\textbf{M} & \textbf{SR} \\\midrule
        SeqBF~\scriptsize{\cite{mou-etal-2016-sequence}} &  4.4 & 0.7 & 0.62 & 0.62 & 7.0 & <100. & 6.9 & 2.1 & 0.52 & 0.53 & 8.7 & <100.\\
        GBS~\scriptsize{\cite{hokamp-liu-2017-lexically}} & 10.1 & 2.8 & 1.49 & 1.50 & 13.5 & $\leq$100. & 13.6 & 4.5 & 1.68 & 1.71 & 15.3 & $\leq$100.\\
        CGMH~\scriptsize{\cite{miao2019cgmh}} & 9.9 & 3.5 & 1.15 & 1.17 & 13.1 & 100. & 12.3 & 4.6 & 1.41 & 1.45 & 14.6 & 100. \\
        POINTER~\scriptsize{\cite{zhang-etal-2020-pointer}} & 8.7 & 1.6 & 2.11 & 2.12 & 14.3 & 100. & 10.6 & 2.4 & 2.14 & 2.16 & 16.8 & 100.\\
        CBART~\scriptsize{\cite{he-2021-parallel}} & 15.6 & 6.6 & 2.16 & 2.19 & 15.2 & 100. & 19.4 & 9.0 & 2.54 & 2.64 & 17.4 & 100. \\
        InstructGPT~\scriptsize{\cite{ouyang2022training}} & 10.1 & 2.8 & 1.72 & 1.73 & 13.0 & 92.33 & 9.3 & 2.4 & 1.42 & 1.44 & 13.6 & 92.17\\
        \midrule
        AutoTemplate & & & & & & & &\\
        \quad w/ T5-small & 16.4 & 6.1 & 3.11 & 3.15 & 15.5 & 100. & 22.5& 9.5 & 3.51 & 3.63 & 17.1 & 100.\\
        \quad w/ T5-base & \underline{18.3} & \underline{7.6} & \underline{3.39} & \underline{3.45} & \underline{16.0} & 100. & \underline{23.7} & \underline{10.8} & \underline{3.62} & \underline{3.76} & \underline{17.8} & 100.\\
        \quad w/ T5-large & \textbf{18.9} & \textbf{8.1} & \textbf{3.49} & \textbf{3.54} & \textbf{16.2} & 100. & \textbf{24.1} & \textbf{11.1} & \textbf{3.68} & \textbf{3.83} & \textbf{17.9} & 100.\\
        \bottomrule
    \end{tabular}
    }
    \caption{Results of keywords-to-sentence generation on the One-Billion-Word and Yelp datasets. \textbf{Bold-faced} and \underline{underlined} denote the best and second-best scores
respectively. Baseline results are copied from \citet{he-2021-parallel}. B2/4 denotes BLEU-2/4, N2/4 denotes NIST-2/4, M denotes METEOR-v1.5, and SR denotes the success rate of lexical constraint satisfaction.
}
    \label{tab:key2text}
\end{table*}

\section{Experiments}
We present experiments across two tasks: keywords-to-sentence generation (\S\ref{sub:k2sg}), and entity-centric summarization (\S\ref{sub:entity_guilded}).

\subsection{Keywords-to-Sentence Generation}
\label{sub:k2sg}
Keywords-to-sentence generation is a task to generate a sentence that includes pre-specified keywords as lexical constraints.
We will show that AutoTemplate is a simple yet effective method to perform this problem without relying on any complex decoding algorithms.

\paragraph{Dataset}
We use One-Billion-Word and the Yelp dataset following the previous studies~\cite{miao2019cgmh,zhang-etal-2020-pointer,he-2021-parallel}. 
One-Billion-Word is a dataset for language modeling based on the WMT 2011 news crawl data~\cite{Chelba2014OneBW}. The Yelp dataset is based on the Yelp open dataset.\footnote{\url{https://www.yelp.com/dataset}} We utilized the publicly available pre-processed dataset,\footnote{\url{https://github.com/NLPCode/CBART}} which consists of 1M, 0.1M sentences for training and development sets, respectively, and 6k sentences with 1-6 pre-specified keywords for test sets, which we summarized in Table~\ref{tab:data-stats}.

\begin{table}[t]
    \centering
    \scriptsize
    \resizebox{\columnwidth}{!}{
    \begin{tabular}{c|ccc}
        \toprule
        \textbf{Data} & \# example & output len. & \# constraints\\\midrule
        1B-Words & 12M & 27.08 & 1 -- 6\\
        Yelp & 13M & 34.26 & 1 -- 6 \\\midrule
        CNNDM &  312k & 70.58 & 4.53\\
        XSum & 226k & 29.39 & 2.11\\
        \bottomrule
    \end{tabular}
    }
    \caption{Dataset Statistics: The output length is the number of BPE tokens per example using the T5 tokenizer. For the summarization datasets, the average number of constraints per example is shown.}
    \label{tab:data-stats}
\end{table}

\paragraph{Baselines}
For the baselines, we used strong competitive models for lexically constrained text generation, including SeqBF~\cite{mou-etal-2016-sequence}, GBS~\cite{hokamp-liu-2017-lexically}, CGMH~\cite{miao2019cgmh}, POINTER~\cite{zhang-etal-2020-pointer}, CBART~\cite{he-2021-parallel}, and InstructGPT~\cite{ouyang2022training}.
SeqBF, GBS, and CGMH are implemented on top of GPT2-small~\cite{radford2019language} (117M parameters). POINTER is implemented on BERT-large~\cite{devlin-etal-2019-bert} (340M parameters), CBART is on BART-large~\cite{lewis-etal-2020-bart} (406M parameters), and InstructGPT has 175B parameters.

\paragraph{Model}
We instantiate the template generation model based on the Transformer~\cite{vaswani2017transformer} initialized with T5 checkpoints~\cite{raffel2020t5} implemented on \texttt{transformers} library~\cite{wolf-etal-2020-transformers}. We specifically utilized the T5-v1.1-small (60M), T5-v1.1-base (220M parameters), and T5-v1.1-Large (770M parameters).
To train the model, we used AdamW optimizer~\cite{loshchilov2018decoupled} with a linear scheduler and warmup, whose initial learning rate is set to 1e-5, and label smoothing~\cite{szegedy2016rethinking} with a label smoothing factor of 0.1.

Since the dataset used in this experiment is a set of raw texts, we randomly select 1 to 6 words from the text and decompose them into constraint lexicons $\mathcal{Z}$ and a template $\tilde{y}$ to create the AutoTemplate training data.
Note that the constraint lexicons $\mathcal{Z}$ were selected from the words excluding punctuations and stopwords~\cite{loper-bird-2002-nltk}.

\paragraph{Metrics}
All performance is measured with the BLEU-2/4~\cite{papineni-etal-2002-bleu}, NIST-2/4 scores~\cite{Doddington2002AutomaticEO}, and METEOR v1.5~\cite{denkowski-lavie-2014-meteor}.
Following the previous study, we show the averaged performance across the number of keywords~\cite{he-2021-parallel}.

\paragraph{Results}
Table~\ref{tab:key2text} shows the results of keywords-to-sentence generation.
First, the performance of GBS and InstructGPT is not as high as non-autoregressive methods. In general, autoregressive decoding produces better text quality than non-autoregressive decoding. However, since GBS is not conditioned on the keywords, it sometimes produces more general text that does not satisfy the keyword constraint. Also, InstructGPT tries to generate sentence according to the instructions, but our experiments show that it frequently fails to include constrained keywords.

Second, among the non-autoregressive baseline models, CBART outperforms CGMH and POINTER.
This suggests that encoder-decoder-based models such as CBART can produce higher-quality text than decoder-only models such as CGMH and POINTER.

Finally, AutoTemplate consistently outperforms all the baselines on both datasets by a large margin while keeping the success rate at 100\% regardless of the model size.
This indicates that AutoTemplate could take advantage of both autoregressive decoding and encoder-decoder models as described above.
We also confirm that using larger T5 models consistently improves text generation quality across all metrics.

Table~\ref{tab:onebillion-1} and \ref{tab:yelp_1} show qualitative examples of generated texts of CBART and AutoTemplate and human written reference.
The examples show that the AutoTemplate generates long and fluent sentences while the CBART tends to generate short text in Table~\ref{tab:onebillion-1} or non-fluent text in Table~\ref{tab:yelp_1}.

\begin{table}[t]
    \centering
    \scriptsize
    \resizebox{\columnwidth}{!}{
    \begin{tabular}{p{7cm}}
        \toprule
        \textbf{Keywords}:\quad\colorhl{ggreen}{leading},\colorhl{oorange}{currency},\colorhl{bblue}{software},\colorhl{ppurple}{industry}\\\midrule
        \textbf{Reference}:\quad Transoft International , Inc. is a \colorhl{ggreen}{leading} provider of \colorhl{oorange}{currency} supply chain management \colorhl{bblue}{software} solutions for the banking \colorhl{ppurple}{industry} .\\\midrule
        \textbf{CBART}:\quad The \colorhl{ggreen}{leading} edge \colorhl{oorange}{currency} trading \colorhl{bblue}{software} \colorhl{ppurple}{industry} .\\\midrule
        \textbf{AutoTemplate}: The company is a \colorhl{ggreen}{leading} provider of \colorhl{oorange}{currency} management \colorhl{bblue}{software} to the financial services \colorhl{ppurple}{industry}.\\
        \bottomrule
    \end{tabular}
    }
    \caption{Example generations for the keywords-to-sentence generation on One-billion-word.}
    \label{tab:onebillion-1}
\end{table}

\begin{table}[t]
    \centering
    \scriptsize
    \resizebox{\columnwidth}{!}{
    \begin{tabular}{p{7cm}}
        \toprule
        \textbf{Keywords}:\quad\colorhl{ggreen}{nail},\colorhl{oorange}{salon},\colorhl{bblue}{always},\colorhl{ppurple}{world} \\\midrule
        \textbf{Reference}:\quad this is the very best \colorhl{ggreen}{nail} \colorhl{oorange}{salon} ! i \colorhl{bblue}{always} see amanda , her workmanship is out of this \colorhl{ppurple}{world} !\\\midrule
        \textbf{CBART}:\quad this is my favorite \colorhl{ggreen}{nail} \colorhl{oorange}{salon} in town ! \colorhl{bblue}{always} clean , friendly and the \colorhl{ppurple}{world} amazing .\\\midrule
        \textbf{AutoTemplate}: I have been going to this \colorhl{ggreen}{nail} \colorhl{oorange}{salon} for over a year now. they \colorhl{bblue}{always} do a great job, and the prices are out of this \colorhl{ppurple}{world}.\\
        \bottomrule
    \end{tabular}
    }
    \caption{Example generations for the keywords-to-sentence generation on Yelp.}
    \label{tab:yelp_1}
\end{table}

\subsection{Entity-guided Summarization}
\label{sub:entity_guilded}
Automatic text summarization distills essential information in a document into short paragraphs, but different readers might want to know different things about specific entities, such as people or places. Thus, one summary might not meet all readers' needs. Entity-guided summarization aims to generate a summary focused on the entities of interest.
This experiment demonstrates that AutoTemplate can produce summaries that satisfy lexical constraints, even under complex entity conditioning.

\begin{table*}[t]
    \centering
    \scriptsize
    \resizebox{2\columnwidth}{!}{
    \begin{tabular}{l|ccccc|ccccc}
    \toprule
    \multirow{2}{*}{\textbf{Model}} & \multicolumn{5}{c|}{\textbf{CNNDM}} & \multicolumn{5}{c}{\textbf{XSum}}\\
        & R1 & R2 & RL & BS & SR & R1 & R2 & RL & BS & SR \\\midrule
        \textit{reported results} & & & & & & & &\\
        \quad BART~\scriptsize{\cite{lewis-etal-2020-bart}} & 44.24 & 21.25 & 41.06 & 0.336 & - & 45.14 & 22.27 & 37.25 & - & - \\
        \quad CTRLSum~\scriptsize{\cite{he-etal-2022-ctrlsum}} & 48.75 & 25.98 & 45.42 & 0.422 & - & - & - & - & - & -\\
        \textit{our implementation} & & & & & & & & \\
        \quad BART~\scriptsize{\cite{lewis-etal-2020-bart}} & 44.20 & 21.28 & 41.02 & 0.358 & 26.12 & 44.21 & 20.93 & 35.18 & 0.510 & 46.69 \\
        \quad CTRLSum~\scriptsize{\cite{he-etal-2022-ctrlsum}} & 47.57 & 25.56 & 44.30 & 0.437 & 75.46 & 50.07 & 26.73 & 40.90 & 0.581 & 86.32\\\midrule
        AutoTemplate &. & & & & & &\\
        \quad w/ T5-base & \underline{51.02} & \underline{27.59} & \underline{47.85} & \underline{0.441} & 100. & \underline{50.49} & \underline{28.19} & \underline{43.89} & \underline{0.591} & 100. \\
        \quad w/ T5-large & \textbf{52.56} & \textbf{29.33} & \textbf{49.38} & \textbf{0.465} & 100. & \textbf{52.65} & \textbf{30.52} & \textbf{46.19} & \textbf{0.614} & 100.\\
    \bottomrule
    \end{tabular}
    }
    \caption{Results of entity-guided summarization with oracle entities on CNNDM and XSum datasets. R1/2/L denotes ROUGE-1/2/L, BS denotes BERTScore, and SR denotes the success rate of lexical constraint satisfaction. \textbf{Bold-faced} and \underline{underlined} denote the best and second-best scores respectively.}
    \label{tab:entity-sum}
\end{table*}

\begin{figure}[t]
    \centering
    \includegraphics[width=\linewidth]{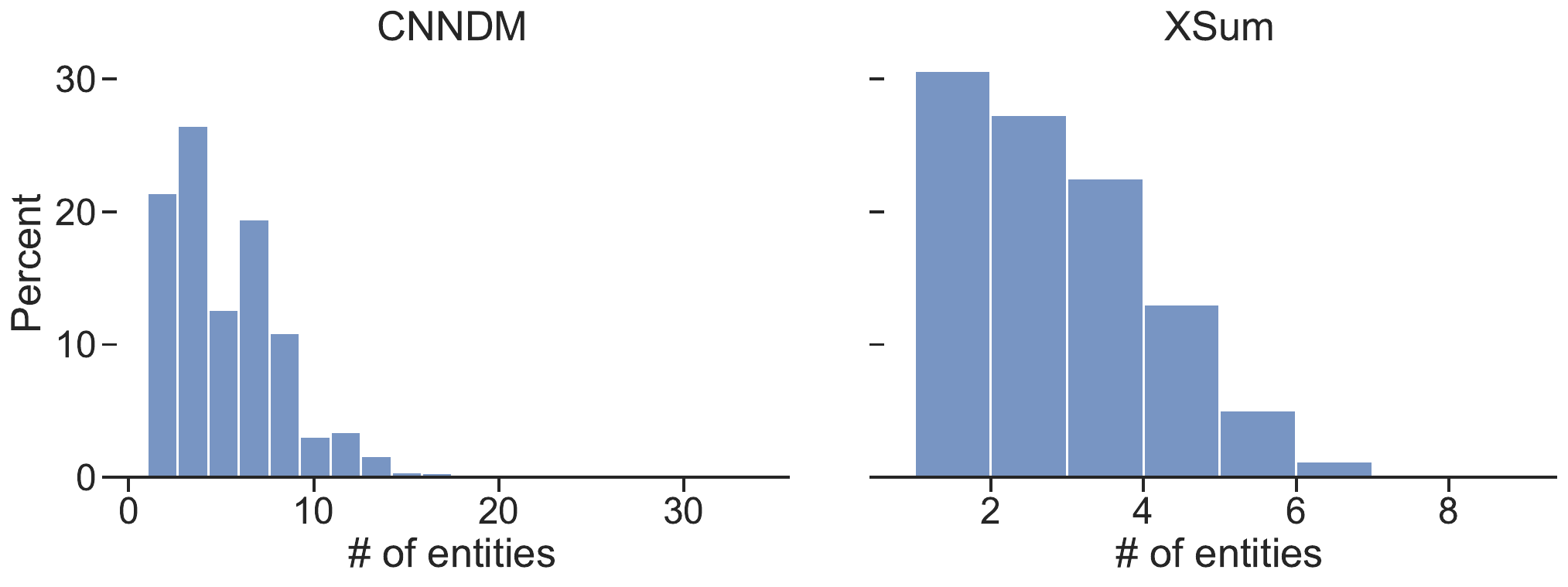}
    \caption{Distribution of the number of oracle entities. The CNNDM dataset (left) tends to have longer summaries and contains more entities than the XSUM dataset. As the number of entities increases, it becomes more and more difficult to include all the entities in the generated summary.}
    \label{fig:entity-dist}
\end{figure}

\paragraph{Dataset}
We use CNNDM dataset~\cite{hermann2015teaching} and XSum dataset~\cite{narayan-etal-2018-dont} for the experiment.
We simulate the entity-guided summarization setting by providing the oracle entity sequence from the gold summary as lexical constraints.
Specifically, we use stanza, an off-the-shelf NER parser~\cite{qi-etal-2020-stanza}, to parse the oracle entity sequence from the gold summary to create entity-guided summarization data.
As summarized in the statistics in Table~\ref{tab:data-stats} and more detailed entity distributions in Figure~\ref{fig:entity-dist}, the CNNDM dataset tends to have more entities than the XSum dataset.
Note that one instance in the test set of the CNNDM dataset has a 676-word reference summary with 84 oracle entities, which is difficult to deal with large pre-trained language models, so we excluded it from the success rate evaluation.

\paragraph{Baselines}
We used competitive models as baselines, including fine-tuned BART~\cite{lewis-etal-2020-bart} and CTRLSum~\cite{he-etal-2022-ctrlsum}. Similar to AutoTemplate, CTRLSum further conditions the input with lexical constraints and generates the output. The difference is that CTRLSum directly generates the output text, while AutoTemplate generates the corresponding template.

\paragraph{Model}
We use the same training configurations to instantiate the model used in the keywords-to-sentence generation task.
To build the training dataset, we use the masked gold summary by the oracle entity sequence as the output template $\tilde{y}$ as described in \S\ref{sec:method}, 
At inference time, we use the oracle entity sequence and the source document as input to generate the template and post-process to produce the output summary.

\paragraph{Metrics}
We evaluate the entity-guided summarization performance using F1 scores of ROUGE-1/2/L~\cite{lin-2004-rouge},\footnote{\url{https://github.com/pltrdy/files2rouge}} BERTScore~\cite{Zhang*2020BERTScore:},\footnote{\url{https://github.com/Tiiiger/bert_score}} and the success rate of entity constraint satisfaction.
Note that our evaluation protocol for the success rate of entity constraint satisfaction is different and more difficult than in previous studies.~\cite{fan-etal-2018-controllable,he-etal-2022-ctrlsum}.
While the previous studies measure whether a \textit{single} specified entity is included in the generated summary, this study measures whether \textit{all} oracle entities are included.

\begin{figure}[t]
    \centering
    \includegraphics[width=\linewidth]{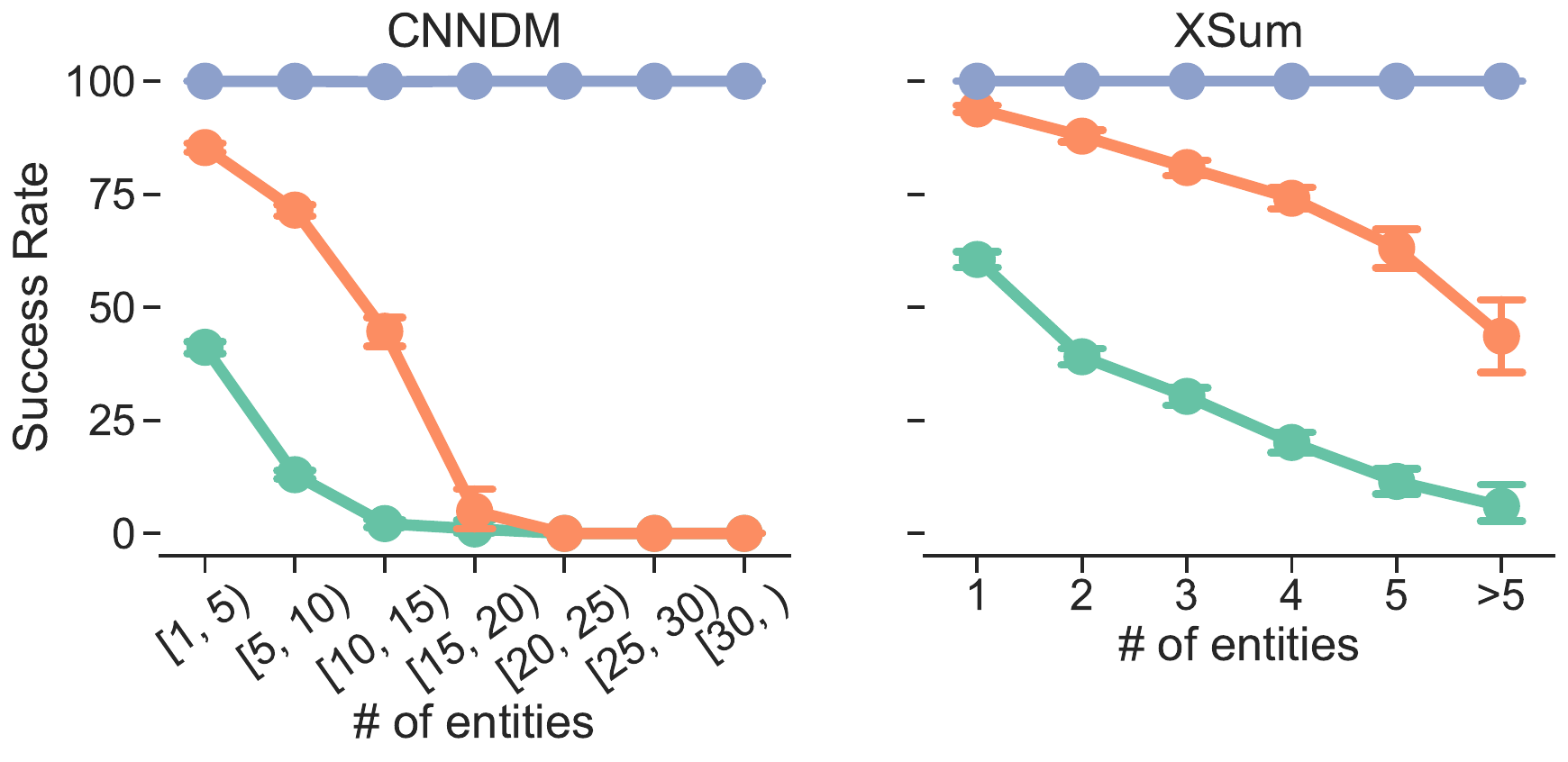}
    \caption{Success rate of entities included in the generated summary at a different number of entities. The \textbf{\textcolor{ggreen}{green line}} denotes the BART model~\cite{lewis-etal-2020-bart}, the \textbf{\textcolor{oorange}{orange line}} denotes the CTRLSum model~\cite{he-etal-2022-ctrlsum}, and \textbf{\textcolor{bblue}{blue line}} denotes AutoTemplate model. These graphs show that CTRLSum can include a limited number of entities in summary with a high chance. However, it becomes more and more difficult as the number of entities increases, while AutoTemplate always satisfies the constraint.}
    \label{fig:success-rate}
\end{figure}
\paragraph{Results}
Table~\ref{tab:entity-sum} shows the results of entity-guided summarization.
CTRLSum and AutoTemplate show improvements in summarization performance compared to the standard BART model, indicating that entity guidance contributes to the improvement in summarization performance.

On the other hand, while AutoTemplate always satisfies entity constraints, CTRLSum shows a constraint satisfaction success rate of 75.46\% for CNNDM and 86.32\% for XSum, characterizing the difference between AutoTemplate and CTRLSum.
As shown in Figure ~\ref{fig:success-rate}, while CTRLSum shows a high success rate when the number of entity constraints is limited, the success rate decreases monotonically as the number of constraints increases.
In contrast, the AutoTemplate showed a 100\% success rate regardless of the number of entity constraints and the highest summarization quality.

Table~\ref{tab:entsum-cnndm} shows the qualitative examples of the generated summaries by CTRLSum and AutoTemplate.
While CTRLSum could only include 10 of the 18 constraint entities in the generated summary, AutoTemplate covered all entities and generated a fluent summary.

We also show the generated summaries with different entity conditioning by AutoTemplate in Table~\ref{tab:controllable}. We confirmed that AutoTemplate can produce summaries with a different focus using different entity conditioning and can also include constraint entities in the generated summary.

\begin{table*}[t]
    \centering
    \scriptsize
    \begin{tabular}{p{\linewidth}}
        \toprule
        \textbf{Constrained Entities: }$\{$\colorhl{ggreen}{Amir Khan},\colorhl{oorange}{Manny Pacquiao},\colorhl{bblue}{Abu Dhabi},\colorhl{ppurple}{UAE}, \colorhl{lgreen}{Khan},\colorhl{yyellow}{Floyd Mayweather Jr},\\
        \colorhl{ggold}{Las Vegas},\colorhl{ggray}{PacMan},\colorhl{ggreen}{Bob Arum},\colorhl{oorange}{UAE},\colorhl{bblue}{\underline{Khan}},\colorhl{ppurple}{\underline{Muslim}},\colorhl{lgreen}{\underline{Brit}},\colorhl{yyellow}{\underline{the Money Man}},\colorhl{ggold}{\underline{PacMan}},\colorhl{ggray}{\underline{Khan}},\colorhl{ggreen}{\underline{Chris Algieri}}, \colorhl{oorange}{\underline{New York}} $\}$\\\midrule  %
        \textbf{CTRLSum}~\cite{he-etal-2022-ctrlsum}: \colorhl{ggreen}{Amir Khan} could face \colorhl{oorange}{Manny Pacquiao} in \colorhl{bblue}{Abu Dhabi}, \colorhl{ppurple}{UAE}. \colorhl{lgreen}{Khan} has been linked with a fight with \colorhl{yyellow}{Floyd Mayweather Jr} in \colorhl{ggold}{Las Vegas}. The \colorhl{ggray}{PacMan}'s promoter \colorhl{ggreen}{Bob Arum} is keen for a fight in the \colorhl{oorange}{UAE}.\\\midrule
        \textbf{AutoTemplate:}  \colorhl{ggreen}{Amir Khan} could face \colorhl{oorange}{Manny Pacquiao} in \colorhl{bblue}{Abu Dhabi}, \colorhl{ppurple}{UAE}. \colorhl{lgreen}{Khan} is preparing to face \colorhl{yyellow}{Floyd Mayweather Jr} in \colorhl{ggold}{Las Vegas} on May 2. \colorhl{ggray}{PacMan}'s vintage promoter \colorhl{ggreen}{Bob Arum} has to hand a treasure trove of an offer for a fight in the \colorhl{oorange}{UAE} this November or December. \colorhl{bblue}{\underline{Khan}} is a hero of the \colorhl{ppurple}{\underline{Muslim}} world, the \colorhl{lgreen}{\underline{Brit}} would be a huge attraction there. Assuming that \colorhl{yyellow}{\underline{the Money Man}} wins his interim bout with \colorhl{ggold}{\underline{PacMan}} next month, all that would appear to stand between him and his long-awaited mega-fight is the outside chance of a re-match. \colorhl{ggray}{\underline{Khan}} is set to fight \colorhl{ggreen}{\underline{Chris Algieri}} in \colorhl{oorange}{\underline{New York}} next month.\\
    \bottomrule
    \end{tabular}
    \caption{Qualitative comparisons between CTRLSum and AutoTemplate. Constraint entities are extracted from the reference summary (oracle entities). \underline{Underlined entities} are missed by the CTRLSum~\cite{he-etal-2022-ctrlsum} while AutoTemplate can incorporate them into the generated summary.}
    \label{tab:entsum-cnndm}
\end{table*}

\begin{table*}[ht]
    \centering
    \scriptsize
    \begin{tabular}{p{\linewidth}}
    \toprule
        \textbf{Constraint Entities: }  $\{$\colorhl{ggreen}{Game Boy},\colorhl{oorange}{Apple},\colorhl{bblue}{Chris Gallizzi}, \colorhl{ppurple}{Nintendo}$\}$ \\\midrule
        \textbf{AutoTemplate}: Case adds iconic \colorhl{ggreen}{Game Boy} buttons to \colorhl{oorange}{Apple} handset. It also lets gamers play their existing cartridges on their handset. Developer \colorhl{bblue}{Chris Gallizzi} said: 'We wanted to create a retro device that can be easily adapted into any modern gamer's arsenal of devices' \colorhl{ppurple}{Nintendo} advised keeping cartridges away from dust, where possible, to avoid gameplay glitches.\\\midrule
        \textbf{Constraint Entities: } $\{$\colorhl{lgreen}{Hyperkin},\colorhl{ppurple}{Nintendo},\colorhl{yyellow}{Game Boy Color},\colorhl{ggold}{Start and Select}$\}$\\\midrule
        \textbf{AutoTemplate}: \colorhl{lgreen}{Hyperkin} has designed a case that adds the iconic directional arrows from the \colorhl{ppurple}{Nintendo} \colorhl{yyellow}{Game Boy Color}. It was originally devised as part of an April Fool's joke, but the popularity and demand for a real product was so high the firm has announced plans to sell it. It will feature an eight-way D-pad, two action buttons, a \colorhl{ggold}{Start and Select} button, and a battery that can be charged through the phone.\\
        \bottomrule
    \end{tabular}
    \caption{Examples of controlled summary generation by changing constraint entities. By conditioning with different entities, the model can generate summaries with different points of interest for the same source article.}
    \label{tab:controllable}
\end{table*}

\begin{table*}[ht]
    \centering
    \resizebox{\textwidth}{!}{
    \begin{tabular}{l|ccccc|ccccc|cccc|cccc}
        \toprule
        & \multicolumn{10}{c|}{\textbf{Keywords-to-Sentence Generation}} & \multicolumn{8}{c}{\textbf{Entity-guided Summarization}}\\
        & \multicolumn{5}{c|}{\textbf{One-Billion-Word}} & \multicolumn{5}{c|}{\textbf{Yelp}} & \multicolumn{4}{c|}{\textbf{CNNDM}} & \multicolumn{4}{c}{\textbf{XSum}}\\
        & B2 & B4 & N2 & N4 & M & B2 & B4 & N2 & N4 & M & R1 & R2 & RL & BS &  R1 & R2 & RL & BS \\\midrule
        AutoTemplate & 18.3 & 7.6 & 3.39 & 3.45 & 16.0 & 23.7 & 10.8 & 3.62 & 3.76 & 17.8 & 51.02 & 27.59 & 47.85 & 0.441 & 50.49 & 28.19 & 43.89 & 0.591\\
        \quad w/ random init & 17.0 & 6.5 & 3.23 & 3.27 & 15.6 & 22.4 & 9.8 & 3.42 & 3.54 & 17.6 & 38.38 & 11.91 & 35.06 & 0.210 & 39.51 & 15.84 & 32.07 & 0.412 \\
        \quad w/ single mask & 16.6 & 5.9 & 3.15 & 3.19 & 15.0 & 15.9 & 5.2 & 2.86 & 2.92 & 13.8 & 48.05 & 24.53 & 44.69 & 0.387 & 45.67 & 23.07 & 39.31 & 0.493 \\
        \bottomrule
    \end{tabular}
    }
    \caption{Ablation studies for keywords-to-sentence generation and entity-guided summarization tasks using T5-base checkpoints. B2/4 denotes BLEU-2/4, N2/4 denotes NIST-2/4, M denotes METEOR-v1.5, R1/2/L denotes ROUGE-1/2/L, and BS denotes BERTScore.}
    \label{tab:ablation}
\end{table*}
\begin{table}[ht]
    \centering
    \scriptsize
    \resizebox{\columnwidth}{!}{
    \begin{tabular}{l|cc}
        \toprule
        \multirow{2}{*}{Fluency (\%)} & \multicolumn{2}{c}{Keywords-to-Sentence} \\
        & One-billion-words & Yelp \\\midrule
        CBART~\tiny{\cite{he-2021-parallel}} & 94.42 & 93.95\\
        InstructGPT~\tiny{\cite{ouyang2022training}} & 96.57 & 96.94 \\
        AutoTemplate & 97.05 & 98.15\\
        Reference & 97.25 & 90.77 \\\midrule
        \multirow{2}{*}{Fluency (\%)} & \multicolumn{2}{c}{Entity-guided summarization}\\
        & CNNDM & XSum \\\midrule
        BART~\tiny{\cite{lewis-etal-2020-bart}} & 96.77 & 98.88\\
        CTRLSum~\tiny{\cite{he-etal-2022-ctrlsum}} & 96.68 & 99.01 \\
        AutoTemplate & 96.38 & 98.91\\
        Reference & 91.55 & 98.73\\
        \bottomrule
    \end{tabular}
    }
    \caption{Results of fluency evaluations by the acceptability classifier trained on CoLA dataset~\cite{warstadt-etal-2019-neural}.}. %
    \label{tab:fluency}
\end{table}

\section{Analysis}
\label{sec:analysis}
\paragraph{Does AutoTemplate generate fluent text?}
AutoTemplate decomposes the lexically constrained text generation task into template generation and lexicalization tasks.
The template generation task aims to produce unnatural text with placeholders, leading to concerns that the final output text will be less fluent than the directly generating natural text.

To this end, we compare the fluency of the output text by AutoTemplate and baselines.
We specifically used the grammatical acceptability classifier based on \texttt{roberta-large} fine-tuned on CoLA dataset~\cite{warstadt-etal-2019-neural} following \citet{krishna-etal-2020-reformulating}\footnote{\url{https://huggingface.co/cointegrated/roberta-large-cola-krishna2020}} and show the micro averaged accuracy of sentence-level grammaticality.\footnote{Although we can also measure fluency using the perplexity of an external language model, it can assign low perplexity to unnatural texts containing common words~\cite{mir-etal-2019-evaluating}. Therefore, we decided to evaluate fluency using the classifier.}

We show the results in Table~\ref{tab:fluency}. For the keywords-to-sentence generation task, AutoTemplate shows better fluency scores than the CBART model, characterizing the differences between CBART and AutoTemplate. While CBART relies on the non-autoregressive models, which leads to non-fluent text generation, AutoTemplate can be implemented on top of autoregressive models. Thus, AutoTemplate can generate more fluent output text.

For the entity-guided summarization task, AutoTemplate shows similar fluency with the state-of-the-art autoregressive text generation models, including BART and CTRLSum, indicating that the AutoTemplate can generate as fluent text as the state-of-the-art direct generation models.

\paragraph{Importance of Pre-training}
To evaluate the importance of T5 pre-training for AutoTemplate, we performed ablation studies using a \textit{randomly} initialized model.
As shown in Table~\ref{tab:ablation}, we confirmed that the model with pre-training significantly improves the quality of generated text in both keywords-to-sentence generation and entity-guided summarization cases.
Note that the keywords-to-sentence generation model with random initialization generally produced better text quality than the baseline model, CBART, confirming the importance of using autoregressive models.

\paragraph{Are unique placeholders needed?}
Throughout this study, we assumed the unique placeholder tokens according to the order of appearance, i.e., \texttt{<X>}, \texttt{<Y>} and \texttt{<Z>}, so we investigate the importance of this design choice.
We show the performance of AutoTemplate with a single type of placeholder token (i.e., \texttt{<X>} for all placeholders in the template $\tilde{y}$) in Table~\ref{tab:ablation}.
We observed a significant drop in the quality of the generated text for both keywords-to-sentence generation and entity-guided summarization tasks, suggesting the importance of using unique placeholder tokens in the template.

\section{Further Related Work}
\paragraph{Template-based Text Generation}
For classical text generation systems, templates were an important building block~\cite{kukich-1983-design,tanaka-ishii-etal-1998-reactive-content,reiter_dale_2000,angeli-etal-2010-simple}.
The advantage of a template-based system is that it can produce faithful text, but it can produce disfluent text if an inappropriate template is selected. Therefore, the current primary approach is to produce fluent text directly from the input using end-to-end neural generation models.

More recent studies have focused mainly on using templates as an auxiliary signal to control the stylistic properties of the output text, such as deriving templates as latent variables~\cite{wiseman-etal-2018-learning,li-rush-2020-posterior,fu2020latent} and using retrieved exemplars as soft templates~\cite{cao-etal-2018-retrieve,peng-etal-2019-text,hossain-etal-2020-simple}.

\paragraph{Copy mechanism}
The copy mechanism was originally introduced to deal with the out-of-vocabulary problem in machine translation by selecting the words from the source for the generation in addition to the vocabulary, such as the unknown word replacement with post-processing~\cite{jean-etal-2015-using,luong-etal-2015-addressing}, and the joint modeling of unknown word probabilities into encoder-decoder models~\cite{gu-etal-2016-incorporating,gulcehre-etal-2016-pointing}, but with the advent of subword units~\cite{sennrich-etal-2016-neural,kudo-2018-subword}, the unknown word problem has been diminished. Thus, the copy mechanism is not widely used now for handling out-of-vocabulary problems.

However, the copy mechanism still plays a vital role in more complex text generation tasks such as involving numerical computation~\cite{murakami-etal-2017-learning,suadaa-etal-2021-towards} or logical reasoning~\cite{chen-etal-2020-logical}.
Specifically, they produce special tokens that serve as placeholders and replace them with the desired words in post-processing.
AutoTemplate adapts a similar copy mechanism to perform lexically constrained text generation, showing that it can cover all the constrained entities in its outputs, even for more complex conditioning (more than ten entities).

\section{Conclusions}
This study proposes AutoTemplate, a simple yet effective framework for lexically constrained text generation. 
The core idea is to decompose lexically constrained text generation into two steps, template generation, and lexicalization, by converting the input and output formats. The template generation can be done with standard encoder-decoder models with beam search so that AutoTemplate can perform lexically constrained text generation without using dedicated decoding algorithms such as non-autoregressive decoding and constrained beam search.
Experimental results show that the AutoTemplate significantly outperforms the competitive baselines across keywords-to-sentence generation and entity-guided summarization tasks while satisfying the lexical constraints.

\section{Limitations}
This study proposes a method to perform hard lexically constrained text generation and shows that our proposed method could generate high-quality text in terms of the automatic evaluation metrics while satisfying the lexical constraints, but this does not guarantee the faithfulness of generated text.
For example, in the summarization task, our method does not directly generate entities prone to errors, so the risk of generating summaries with unfaithful entities to the input text could be lower than existing methods. Still, the risk of generating unfaithful text in other areas remains.
For the evaluation, we didn't have LLM-as-a-judge due to the budget constraint even though it shows a high correlation with human judgment~\cite{liu-etal-2023-g,wu-etal-2024-less}.

\bibliography{custom}

\end{document}


\maketitle
\appendix

\section{More qualitative examples}
\label{sec:k2s-more}
Table~\ref{tab:onebillion_2}-\ref{tab:yelp_3} show more qualitative examples of keywords-to-sentence generation task.
\begin{table}[t]
    \centering
    \footnotesize
    \begin{tabular}{p{7cm}}
        \toprule
        \textbf{Keywords}: \colorhl{ggreen}{government},\colorhl{oorange}{ability},\colorhl{bblue}{companies},\colorhl{ppurple}{legal} \\\midrule
        \textbf{Reference}: Generally , the \colorhl{ggreen}{government} has the \colorhl{oorange}{ability} to compel the cooperation of private \colorhl{bblue}{companies} and assure them \colorhl{ppurple}{legal} immunity with a valid court order .\\\midrule
        \textbf{CBART}: The \colorhl{ggreen}{government} has restricted the \colorhl{oorange}{ability} of insurance \colorhl{bblue}{companies} to take \colorhl{ppurple}{legal} action .\\\midrule
        \textbf{AutoTemplate:} The \colorhl{ggreen}{government} has the \colorhl{oorange}{ability} to force \colorhl{bblue}{companies} to comply with \colorhl{ppurple}{legal} requirements, he said.\\
        \bottomrule
    \end{tabular}
    \caption{Example generations for the keywords-to-sentence generation on One-billion-word.}
    \label{tab:onebillion_2}
\end{table}

\begin{table}[t]
    \centering
    \footnotesize
    \begin{tabular}{p{7cm}}
        \toprule
        \textbf{Keywords}:  \colorhl{ggreen}{time},\colorhl{oorange}{voters},\colorhl{bblue}{primary},\colorhl{ppurple}{days} \\\midrule
        \textbf{Reference}: At the same \colorhl{ggreen}{time} , he said the more he appears before \colorhl{oorange}{voters} , the better he does on \colorhl{bblue}{primary} \colorhl{ppurple}{days} .\\\midrule
        \textbf{CBART}: The last \colorhl{ggreen}{time} , the \colorhl{oorange}{voters} were in the \colorhl{bblue}{primary} , two \colorhl{ppurple}{days} before Nov .\\\midrule
        \textbf{AutoTemplate}: At the same \colorhl{ggreen}{time}, \colorhl{oorange}{voters} will be able to cast their ballots during the \colorhl{bblue}{primary} \colorhl{ppurple}{days}, he said.\\
        \bottomrule
    \end{tabular}
    \caption{Example generations for the keywords-to-sentence generation on One-billion-word.}
    \label{tab:onebillion_3}
\end{table}

\begin{table}[t]
    \centering
    \footnotesize
    \begin{tabular}{p{7cm}}
        \toprule
        \textbf{Keywords}:  \colorhl{ggreen}{experience}, \colorhl{oorange}{top}, \colorhl{bblue}{easily}, \colorhl{ppurple}{driver} \\\midrule
        \textbf{Reference}: my \colorhl{ggreen}{experience} with lv cans was \colorhl{oorange}{top} notch . cab was \colorhl{bblue}{easily} flagged just off the strip , the route was direct and the \colorhl{ppurple}{driver} was very nice .\\\midrule
        \textbf{CBART}: the whole \colorhl{ggreen}{experience} was \colorhl{oorange}{top} notch , \colorhl{bblue}{easily} by the \colorhl{ppurple}{driver} .\\\midrule
        \textbf{AutoTemplate:} i had a great \colorhl{ggreen}{experience} with this company. they were on \colorhl{oorange}{top} of everything. i was \colorhl{bblue}{easily} able to get a \colorhl{ppurple}{driver} to pick me up at my hotel.\\
        \bottomrule
    \end{tabular}
    \caption{Example generations for the keywords-to-sentence generation on Yelp.}
    \label{tab:yelp_2}
\end{table}

\begin{table}[t]
    \centering
    \footnotesize
    \begin{tabular}{p{7cm}}
        \toprule
        \textbf{Keywords}:  \colorhl{ggreen}{southern}, \colorhl{oorange}{fresh}, \colorhl{bblue}{made}, \colorhl{ppurple}{friendly} \\\midrule
        \textbf{Reference}: absolutely , the best pizza in \colorhl{ggreen}{southern} nevada ! the pizza is always \colorhl{oorange}{fresh} , \colorhl{bblue}{made} fresh as ordered . the wait staff is very \colorhl{ppurple}{friendly} and effecient !\\\midrule
        \textbf{CBART}: great \colorhl{ggreen}{southern} food , \colorhl{oorange}{fresh} and \colorhl{bblue}{made} with \colorhl{ppurple}{friendly} staff .\\\midrule
        \textbf{AutoTemplate}: this is the best \colorhl{ggreen}{southern} food i have ever had. everything is \colorhl{oorange}{fresh} and \colorhl{bblue}{made} to order. the staff is very \colorhl{ppurple}{friendly} and helpful. i will definitely be back.\\
        \bottomrule
    \end{tabular}
    \caption{Example generations for the keywords-to-sentence generation on Yelp.}
    \label{tab:yelp_3}
\end{table}

\section{Additional Experimental Details}

\subsection{Training details}
Major hyper-parameters for training models are reported in Table \ref{tab:autotemplate_param} following the "Show-You-Work" style suggested by \citet{dodge-etal-2019-show}.

\begin{table*}[ht]
    \centering
    \small
    \begin{tabular}{cc}
        \toprule
       \textbf{Computing infrastructure} & NVIDIA A100\\
       \midrule
       \textbf{Training duration} & 4h \\
       \midrule
       \textbf{Search strategy} & Manual tuning \\\midrule
       \textbf{Model implementation} & \url{[MASK]}\\
       \midrule
       \textbf{Model checkpoint} & \url{[MASK]}\\
       \bottomrule
    \end{tabular}

    \vspace{3mm}
    \resizebox{2\columnwidth}{!}{
    \begin{tabular}{ccc}
    \toprule
    \textbf{Hyperparameter} & \textbf{Search space} & \textbf{Best assignment} \\
    \midrule
    \# of training steps & 50,000 & 50,000\\ 
    \midrule
    validation interval & 5,000 & 5,000\\
    \midrule 
    batch size & 32 & 32\\
    \midrule
    initial checkpoint for small models & \texttt{google/t5-v1\_1-small} & \texttt{google/t5-v1\_1-small} \\
    initial checkpoint for base models & \texttt{google/t5-v1\_1-base} & \texttt{google/t5-v1\_1-base} \\
    initial checkpoint for large models & \texttt{google/t5-v1\_1-large} & \texttt{google/t5-v1\_1-large} \\
    \midrule
    label-smoothing~\tiny{\cite{szegedy2016rethinking}} & \emph{choice}[0.0, 0.1] & 0.1 \\
    \midrule
    learning rate scheduler & linear schedule with warmup & linear schedule with warmup\\
    \midrule
    warmup steps & 5,000 & 5,000 \\
    \midrule
    learning rate optimizer & AdamW~\tiny{\cite{loshchilov2019decoupled}} & AdamW~\tiny{\cite{loshchilov2019decoupled}}\\
    \midrule
    AdamW $\beta_1$ & 0.9 & 0.9\\
    \midrule
    AdamW $\beta_2$ & 0.999 & 0.999\\
    \midrule
    learning rate & 5e-5 & 5e-5 \\
    \midrule
    weight decay & \emph{choice}[0.0, 1e-3, 1e-2] & 1e-2 \\
    \midrule
    max grad norm & 0.1 & 0.1 \\
    \midrule
    beam width for keywords-to-sentence & 4 & 4 \\
    beam width for entity-guided summarization on CNNDM & 8 & 8\\
    beam width for entity-guided summarization on XSUM & 6 & 6\\
    \bottomrule
    \end{tabular}
    }
    \caption{AutoTemplate search space and the best assignments.}
    \label{tab:autotemplate_param}
\end{table*}

\section{Experimental details of InstructGPT}
\label{sec:gpt3}
We empirically evaluated the zero-shot capability of InstructGPT~\cite{ouyang2022training} for keywords-to-sentence generation task. We specifically used \texttt{text-davinci-003} checkpoint and the prompt: \texttt{"Please create a sentence that must contain the following keywords: \{\{', '.join(keywords)\}\}."} to generate sentences that includes the pre-specified keywords. To obtain deterministic output text, we use the temperature parameter 0.

\section{Full results of keywords-to-sentence generation}
We show non-aggregated results of keywords-to-sentence generation in Table~\ref{tab:key2text_all}. The results show that the AutoTemplate consistently outperforms baseline models.
\begin{table*}[t]
    \centering
    \resizebox{\textwidth}{!}{
    \footnotesize
    \begin{tabular}{l|cccccc|cccccc}
        \toprule
        \multirow{2}{*}{\textbf{\# of keywords = 1}} & \multicolumn{6}{c}{\textbf{One-Billion-Word}} & \multicolumn{6}{c}{\textbf{Yelp}}\\
        & \textbf{B2} & \textbf{B4} & \textbf{N2} & \textbf{N4} & \textbf{M} & \textbf{SR} & \textbf{B2} & \textbf{B4} & \textbf{N2} & \textbf{N4} &\textbf{M} & \textbf{SR} \\\midrule
        CBART~\scriptsize{\cite{he-2021-parallel}} & 3.81 & 0.61 & 0.34 & 0.34 & 6.77 & 100. & 5.71 & 1.66 & 0.31 & 0.32 & 8.33 & 100. \\
        InstructGPT~\scriptsize{\cite{ouyang2022training}} & 2.49 & 0.32 & 0.24 & 0.24 & 5.93 & 98.4 & 2.39 & 0.31 & 0.18 & 0.18 & 6.34 & 98.5\\
        \midrule
        AutoTemplate & & & & & & & &\\
        \quad w/ T5-small & 5.56 & 0.88 & 1.23 & 1.23 & 9.04 & 100. & 9.80 & 2.46 & 1.65 & 1.68 & 10.84 & 100.\\
        \quad w/ T5-base & \underline{6.01} & \underline{1.01} & \underline{1.36} & \underline{1.36} & \underline{8.82} & 100. & \underline{9.95} & \underline{2.52} & \underline{1.68} & \underline{1.68} & \underline{10.94} & 100.\\
        \quad w/ T5-large & \textbf{6.19} & \textbf{1.16} & \textbf{1.40} & \textbf{1.40} & \textbf{8.74} & 100. & \textbf{9.78} & \textbf{2.44} & \textbf{1.67} & \textbf{1.69} & \textbf{10.99} & 100.\\
        \midrule
        \textbf{\# of keywords = 2} & \textbf{B2} & \textbf{B4} & \textbf{N2} & \textbf{N4} & \textbf{M} & \textbf{SR} & \textbf{B2} & \textbf{B4} & \textbf{N2} & \textbf{N4} &\textbf{M} & \textbf{SR} \\\midrule
        CBART~\scriptsize{\cite{he-2021-parallel}} & 7.25 & 1.91 & 0.68 & 0.68 & 10.02 & 100. & 9.67 & 3.14 & 0.74 & 0.76 & 11.75 & 100. \\
        InstructGPT~\scriptsize{\cite{ouyang2022training}} & 4.57 & 0.84 & 0.48 & 0.49 & 8.68 & 95.2 & 3.94 & 0.66 & 0.30 & 0.30 & 8.89 & 95.0\\
        \midrule
        AutoTemplate & & & & & & & &\\
        \quad w/ T5-small & 8.23 & 1.77 & 1.72 & 1.73 & 11.49 & 100. & 13.46 & 3.94 & 2.14 & 2.18 & 13.09 & 100.\\
        \quad w/ T5-base & \underline{9.76} & \underline{2.52} & \underline{2.00} & \underline{2.02} & \underline{11.39} & 100. & \underline{13.71} & \underline{4.16} & \underline{2.18} & \underline{2.22} & \underline{13.36} & 100.\\
        \quad w/ T5-large & \textbf{10.06} & \textbf{2.59} & \textbf{2.05} & \textbf{2.06} & \textbf{11.35} & 100. & \textbf{13.55} & \textbf{4.04} & \textbf{2.17} & \textbf{2.21} & \textbf{13.25} & 100.\\
        \midrule
        \textbf{\# of keywords = 3} & \textbf{B2} & \textbf{B4} & \textbf{N2} & \textbf{N4} & \textbf{M} & \textbf{SR} & \textbf{B2} & \textbf{B4} & \textbf{N2} & \textbf{N4} &\textbf{M} & \textbf{SR} \\\midrule
        CBART~\scriptsize{\cite{he-2021-parallel}} & 11.68 & 3.84 & 1.26 & 1.27 & 13.30 & 100. & 16.03 & 6.48 & 1.73 & 1.77 & 15.75 & 100. \\
        InstructGPT~\scriptsize{\cite{ouyang2022training}} & 7.58 & 1.58 & 0.97 & 0.97 & 11.52 & 92.5 & 6.67 & 1.30 & 0.66 & 0.67 & 11.95 & 92.2\\
        \midrule
        AutoTemplate & & & & & & & &\\
        \quad w/ T5-small & 13.20 & 3.73 & 2.60 & 2.62 & 13.76 & 100. & 19.17 & 7.09 & 2.99 & 3.07 & 15.66 & 100.\\
        \quad w/ T5-base & \underline{15.26} & \underline{5.13} & \underline{2.85} & \underline{2.88} & \underline{14.08} & 100. & \underline{19.82} & \underline{7.81} & \underline{3.05} & \underline{3.15} & \underline{16.20} & 100.\\
        \quad w/ T5-large & \textbf{16.05} & \textbf{5.53} & \textbf{3.00} & \textbf{3.03} & \textbf{14.26} & 100. & \textbf{20.20} & \textbf{8.11} & \textbf{3.09} & \textbf{3.19} & \textbf{16.01} & 100.\\
        \midrule
        \textbf{\# of keywords = 4} & \textbf{B2} & \textbf{B4} & \textbf{N2} & \textbf{N4} & \textbf{M} & \textbf{SR} & \textbf{B2} & \textbf{B4} & \textbf{N2} & \textbf{N4} &\textbf{M} & \textbf{SR} \\\midrule
        CBART~\scriptsize{\cite{he-2021-parallel}} & 17.67 & 7.07 & 2.31 & 2.34 & 16.92 & 100. & 22.45 & 10.28 & 3.00 & 3.10 & 19.39 & 100. \\
        InstructGPT~\scriptsize{\cite{ouyang2022training}} & 11.29 & 3.09 & 1.81 & 1.82 & 14.52 & 91.6 & 10.35 & 2.68 & 1.46 & 1.48 & 15.19 & 90.1\\
        \midrule
        AutoTemplate & & & & & & & &\\
        \quad w/ T5-small & 19.04 & 6.54 & 3.76 & 3.81 & 16.51 & 100. & 25.84 & 10.77 & 3.96 & 4.10 & 18.30 & 100.\\
        \quad w/ T5-base & \underline{20.92} & \underline{8.05} & \underline{3.97} & \underline{4.02} & \underline{17.19} & 100. & \underline{26.87} & \underline{12.26} & \underline{4.02} & \underline{4.21} & \underline{19.03} & 100.\\
        \quad w/ T5-large & \textbf{21.23} & \textbf{8.58} & \textbf{4.01} & \textbf{4.08} & \textbf{17.29} & 100. & \textbf{28.04} & \textbf{12.95} & \textbf{4.20} & \textbf{4.36} & \textbf{19.25} & 100.\\
        \midrule
        \textbf{\# of keywords = 5} & \textbf{B2} & \textbf{B4} & \textbf{N2} & \textbf{N4} & \textbf{M} & \textbf{SR} & \textbf{B2} & \textbf{B4} & \textbf{N2} & \textbf{N4} &\textbf{M} & \textbf{SR} \\\midrule
        CBART~\scriptsize{\cite{he-2021-parallel}} & 23.51 & 10.78 & 3.50 & 3.56 & 20.36 & 100. & 27.97 & 13.80 & 4.12 & 4.28 & 22.73 & 100. \\
        InstructGPT~\scriptsize{\cite{ouyang2022training}} & 15.32 & 4.46 & 2.86 & 2.88 & 17.43 & 89.9 & 13.97 & 3.92 & 2.41 & 2.44 & 18.05 & 90.9\\
        \midrule
        AutoTemplate & & & & & & & &\\
        \quad w/ T5-small & 23.47 & 9.76 & 4.33 & 4.40 & 19.58 & 100. & 30.43 & 13.87 & 4.78 & 4.97 & 20.92 & 100.\\
        \quad w/ T5-base & \underline{25.97} & \underline{12.03} & \underline{4.68} & \underline{4.78} & \underline{20.44} & 100. & \underline{32.85} & \underline{16.40} & \underline{4.94} & \underline{5.16} & \underline{22.01} & 100.\\
        \quad w/ T5-large & \textbf{26.89} & \textbf{12.74} & \textbf{4.79} & \textbf{4.89} & \textbf{20.93} & 100. & \textbf{33.11} & \textbf{16.71} & \textbf{5.05} & \textbf{5.28} & \textbf{22.18} & 100.\\
        \midrule
        \textbf{\# of keywords = 6} & \textbf{B2} & \textbf{B4} & \textbf{N2} & \textbf{N4} & \textbf{M} & \textbf{SR} & \textbf{B2} & \textbf{B4} & \textbf{N2} & \textbf{N4} &\textbf{M} & \textbf{SR} \\\midrule
        CBART~\scriptsize{\cite{he-2021-parallel}} & 29.93 & 15.38 & 4.83 & 4.93 & 23.72 & 100. & 34.50 & 18.56 & 5.35 & 5.59 & 26.33 & 100. \\
        InstructGPT~\scriptsize{\cite{ouyang2022training}} & 19.50 & 6.71 & 3.93 & 3.97 & 20.20 & 86.4 & 18.33 & 5.76 & 3.50 & 3.55 & 21.01 & 86.3\\
        \midrule
        AutoTemplate & & & & & & & &\\
        \quad w/ T5-small & 28.69 & 13.79 & 5.00 & 5.10 & 22.87 & 100. & 36.31 & 18.99 & 5.53 & 5.80 & 24.03 & 100.\\
        \quad w/ T5-base & \underline{31.98} & \underline{17.08} & \underline{5.50} & \underline{5.63} & \underline{23.97} & 100. & \underline{38.85} & \underline{21.73} & \underline{5.80} & \underline{6.10} & \underline{25.36} & 100.\\
        \quad w/ T5-large & \textbf{33.20} & \textbf{18.18} & \textbf{5.66} & \textbf{5.80} & \textbf{24.42} & 100. & \textbf{39.63} & \textbf{22.60} & \textbf{5.92} & \textbf{6.24} & \textbf{25.69} & 100.\\
        \bottomrule
    \end{tabular}
    }
    \caption{Comprehensive results of keywords-to-sentence generation on the One-Billion-Word and Yelp datasets. \textbf{Bold-faced} and \underline{underlined} denote the best and second-best scores
respectively. Baseline results are copied from \citet{he-2021-parallel}. B2/4 denotes BLEU-2/4, N2/4 denotes NIST-2/4, M denotes METEOR-v1.5, and SR denotes the success rate of lexical constraint satisfaction.}
    \label{tab:key2text_all}
\end{table*}

\begin{table*}[t]
    \centering
    \footnotesize
    \begin{tabular}{p{\linewidth}}
        \toprule
        \textbf{Constrained Entities: }$\{$\colorhl{ggreen}{Amir Khan},\colorhl{oorange}{Manny Pacquiao},\colorhl{bblue}{Abu Dhabi},\colorhl{ppurple}{UAE}, \colorhl{lgreen}{Khan},\colorhl{yyellow}{Floyd Mayweather Jr},\\
        \colorhl{ggold}{Las Vegas},\colorhl{ggray}{PacMan},\colorhl{ggreen}{Bob Arum},\colorhl{oorange}{UAE},\colorhl{bblue}{\underline{Khan}},\colorhl{ppurple}{\underline{Muslim}},\colorhl{lgreen}{\underline{Brit}},\colorhl{yyellow}{\underline{the Money Man}},\colorhl{ggold}{\underline{PacMan}},\colorhl{ggray}{\underline{Khan}},\colorhl{ggreen}{\underline{Chris Algieri}}, \colorhl{oorange}{\underline{New York}} $\}$\\\midrule  %
        \textbf{BART}~\cite{lewis-etal-2020-bart}: \colorhl{ggreen}{Amir Khan} has been linked with a fight with \colorhl{oorange}{Manny Pacquiao}. The fight could take place in \colorhl{bblue}{Abu Dhabi} in November or December. \colorhl{ggray}{Khan} is preparing to fight \colorhl{ggreen}{Chris Algieri} in \colorhl{oorange}{New York} next month. Pacquiao is preparing to face Floyd Mayweather on May 2 in \colorhl{ggold}{Las Vegas}.\\\midrule
        \textbf{CTRLSum}~\cite{he-etal-2022-ctrlsum}: \colorhl{ggreen}{Amir Khan} could face \colorhl{oorange}{Manny Pacquiao} in \colorhl{bblue}{Abu Dhabi}, \colorhl{ppurple}{UAE}. \colorhl{lgreen}{Khan} has been linked with a fight with \colorhl{yyellow}{Floyd Mayweather Jr} in \colorhl{ggold}{Las Vegas}. The \colorhl{ggray}{PacMan}'s promoter \colorhl{ggreen}{Bob Arum} is keen for a fight in the \colorhl{oorange}{UAE}.\\\midrule
        \textbf{AutoTemplate:}  \colorhl{ggreen}{Amir Khan} could face \colorhl{oorange}{Manny Pacquiao} in \colorhl{bblue}{Abu Dhabi}, \colorhl{ppurple}{UAE}. \colorhl{lgreen}{Khan} is preparing to face \colorhl{yyellow}{Floyd Mayweather Jr} in \colorhl{ggold}{Las Vegas} on May 2. \colorhl{ggray}{PacMan}'s vintage promoter \colorhl{ggreen}{Bob Arum} has to hand a treasure trove of an offer for a fight in the \colorhl{oorange}{UAE} this November or December. \colorhl{bblue}{\underline{Khan}} is a hero of the \colorhl{ppurple}{\underline{Muslim}} world, the \colorhl{lgreen}{\underline{Brit}} would be a huge attraction there. Assuming that \colorhl{yyellow}{\underline{the Money Man}} wins his interim bout with \colorhl{ggold}{\underline{PacMan}} next month, all that would appear to stand between him and his long-awaited mega-fight is the outside chance of a re-match. \colorhl{ggray}{\underline{Khan}} is set to fight \colorhl{ggreen}{\underline{Chris Algieri}} in \colorhl{oorange}{\underline{New York}} next month.\\
        \midrule
        \textbf{Reference}: \colorhl{ggreen}{Amir Khan} could be set to face \colorhl{oorange}{Manny Pacquiao} in \colorhl{bblue}{Abu Dhabi}, \colorhl{ppurple}{UAE}. \colorhl{lgreen}{Khan}'s hopes of taking on \colorhl{yyellow}{Floyd Mayweather Jr} in \colorhl{ggold}{Las Vegas} have faded. \colorhl{ggray}{PacMan}'s promoter \colorhl{ggreen}{Bob Arum} has a mega offer for a \colorhl{oorange}{UAE} fight late in 2015. \colorhl{bblue}{\underline{Khan}} is a hero of the \colorhl{ppurple}{\underline{Muslim}} world and his lure in the Middle East is clear. The \colorhl{lgreen}{\underline{Brit}} will be ringside when \colorhl{yyellow}{\underline{the Money Man}} fights the \colorhl{ggold}{\underline{PacMan}} on May 2. \colorhl{ggray}{\underline{Khan}} must first win interim bout with \colorhl{ggreen}{\underline{Chris Algieri}} in \colorhl{oorange}{\underline{New York}} on May 29.\\
    \bottomrule
    \end{tabular}
    \caption{Qualitative comparisons between CTRLSum and AutoTemplate. Constraint entities are extracted from the reference summary (oracle entities). \underline{Underlined entities} are missed by the CTRLSum~\cite{he-etal-2022-ctrlsum} while AutoTemplate can incorporate them into the generated summary.}
    \label{tab:entsum-cnndm_full}
\end{table*}
\bibliography{custom}